# VQT-Light: Lightweight HDR Illumination Map Prediction with Richer Texture

Xie Kunliang[1]

(1 *College of Electronic Science, National University of Defense Technology, Changsha*  410003)

**Abstract:** Accurate lighting estimation is a significant yet challenging task in computer vision and graphics. However, existing methods either struggle to restore detailed textures of illumination map, or face challenges in running speed and texture fidelity. To tackle this problem, we propose a novel framework (VQT-Light) based on VQVAE and ViT architecture. VQT-Light includes two modules: feature extraction and lighting estimation. First, we take advantages of VQVAE to extract discrete features of illumination map rather than continuous features to avoid "posterior collapse". Second, we capture global context and dependencies of input image through ViT rather than CNNs to improve the prediction of illumination outside the field of view. Combining the above two modules, we formulate the lighting estimation as a multiclass classification task, which plays a key role in our pipeline. As a result, our model predicts light map with richer texture and better fidelity while keeping lightweight and fast. VQT-Light achieves an inference speed of 40FPS and improves multiple evaluation metrics. Qualitative and quantitative experiments demonstrate that the proposed method realizes superior results compared to existing state-of-the-art methods. Code is available at https://github.com/heartxkl/vqt-light.

## 1  Introduction

Lighting estimation has a wide range of practical applications, such as virtual object inserting in augmented reality. However, it is an under-constrained problem as it aims to recover the global environment illumination from a limited field-of-view (FOV) image. As we all know, the process of image formation involves many factors, such as lighting conditions, surface materials, scene geometry, and camera parameters. Therefore, the ill-posed inverse process for estimating the original lighting conditions from an image is quite challenging. It is even more complicated to infer a high-dynamic-range (HDR) panoramic illumination map when the input image is recorded in low-dynamic-range (LDR) with a limited FOV.

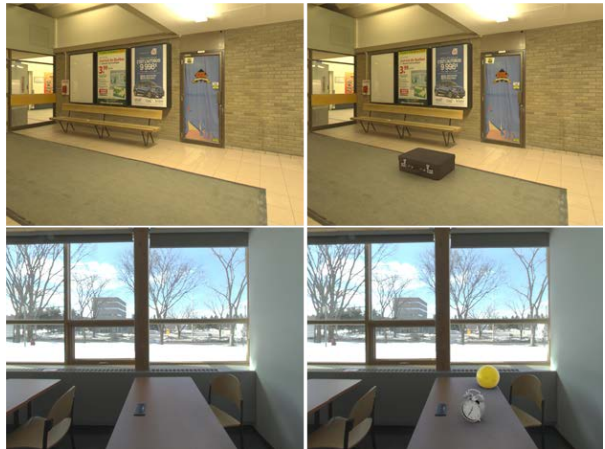

**Fig. 1**: Given a single limited field-of-view image (left), our method predicts the global environment lighting represented as HDR illumination map. Using the predicted HDR panoramas, virtual objects can be realistically relit and composited into images (right).

Thanks to the development of deep learning, lighting estimation has been tackled by regression of lighting parametric models, such as spherical harmonics (SH) [10, 11] and spherical Gaussians (SG) [12, 13, 17, 19], or by direct generation of HDR illumination map [14-23]. However, regression-based methods run quickly but struggle to recover lighting with full-frequency, especially many texture details of light map are lost. Generation-based methods are able to generate plausible textures, but the generated textures may be quite different from the ground truth, and the running speed of these methods is usually slow. The shortcomings of these two kinds of methods have a negative impact on practical application. Thus, there still remains a large space for further exploration in this field.

Our previous work [24] has made a progress in the projection scheme of illumination map, in which we introduce a square equal-area projection (SP) [27] to replace the equirectangular projection (EP), so as to decrease the distortions of angular and area for illumination map. While this point helps to improve the running speed and the accuracy of lighting estimation, the predicted results still lack many texture details leading to pool visual effects when rendering glossy objects.

In this work, we design a novel two-stage framework (VQT-Light) to predict light map with richer texture. In the first stage, we extract discrete features from HDR illumination map based on VQVAE architecture [25]. The discrete features are represented as discrete embedding vectors. In the second stage, taking a single image as input, a lightweight vision transformer (ViT) [26] is used to regress the indices of discrete embedding vectors from the first stage. These indices serve as classification labels. Then the ViT network outputs the predicted indices, which can be indexed to get the predicted embedding vectors from the codebook of VQVAE. As a result, the predicted embeddings are feed into the frozen VQVAE decoder to produce an HDR illumination map. Comparing with [24], our method is able to restore more texture details, resulting in more accurate illumination prediction. As depicted in Figure 1, using the predicted HDR panoramas of our method, virtual objects can be realistically relit and composited into images. Our main contributions are:

1. We design a novel framework (VQT-Light), which combines the advantages of VQVAE and ViT, to predict HDR illumination map with richer texture, and seek for the most suitable structure of VQT-Light for performance trade-offs.
2. We translate the lighting estimation as a multiclass classification task to achieve better performance, to the best of our knowledge, this is the first work that estimates illumination in this manner.
3. Extensive qualitative and quantitative experiments demonstrate the superiority and the effectiveness of our method over existing state-of-the-art methods.

## 2 Related Works

Lighting estimation is a classic problem and has a long history of research in computer vision and graphics. It can be generally divided into traditional techniques and learning-based techniques.

**Traditional Techniques** typically require auxiliary equipment, user intervention or the assumption of known scene geometry, material properties, etc. For example, Debevec'98 [1], Reinhard'04 [2], Reinhard'10 [3] and Debevec'12 [4] place a ball of known material in the scene to estimate an HDR illumination map. Karsch'11 [5] requires user annotations for initial lighting and geometry estimation to recover parametric 3D lighting. Xing'13 [6] predicts spherical harmonics illumination with the help of user annotations about coarse structure and material of the scene. Zhang'16 [7] infers lighting and material with a group of RGBD multi-view images. Maier'17 [8] recover spatially-varying lighting with the assumption of known depth information of scene. Traditional techniques often involve complicated operation, which are not suitable for use by non-professionals.

**Learning-based Techniques** aim to predict illumination automatically through a single or multiple images without any user intervention and auxiliary equipment. These methods can also be divided into regression-based methods and generation-based methods. Regression-based methods choose effective parametric representations of lighting, the corresponding parameters are used as labels to in a regression task, the common lighting parametric representations include SH, SG, wavelets, etc. For example, Cheng'18 [10] regresses the SH parameters of scene lighting with a render loss, which inputs a pair of images taken from the front and rear cameras of a smartphone. Garon'19 [11] predicts lighting in real-time by regressing SH parameters from a background image and local patch. Garnder'19 [12] regresses the intensities, colors, and positions of light sources, then converts the above data into SG functions for illumination estimation. Li'19 [13] regresses the SG parameters of scene lighting with a render loss. Li'20 [17] regresses SG parameters to estimate per-pixel illumination with a cascade network. Bai'22 [19] regresses SG parameters of lighting by a graph neural network. Zhan'21 [18] utilizes Needlets to represent illumination map and regresses sparse coefficients of Needlets with a spherical transport loss. Xie'24 [24] regresses latent codes of illumination map from a single image, the latent codes are extracted by an AutoEncoder network. Unlike regression-based methods, generation-based methods mainly generate illumination map directly instead of regressing light parameters. For example, Song'19 [14] generates illumination map by completing sub-tasks including geometry estimation, scene completion, and LDR-to-HDR estimation. Lighthouse [15] and Wang'21 [16] propose a 3D volumetric model to generate illumination map by means of volume rendering. EM-Light [18] predicts SG parameters of lighting firstly, then generates illumination map using a GAN network with the guidance of the predicted SG parameters. Xu'22 [20] uses both SH and SG to represent illumination map, then project SH and SG parameters into a guidance map, which guides the generation of illumination map in a rendering-aware network. StyleLight [22] proposes a StyleGAN-based network [21] to generate illumination map, which also supports lighting editing. EverLight [23] proposes a lighting co-modulation technique to generate and edit illumination map.

The aforementioned works either struggle to recover texture details of illumination map, or have problems in running speed and texture fidelity. In contrast, we propose a lightweight framework that makes the predicted textures richer and as close to the ground truth as possible while maintaining real-time running speed.

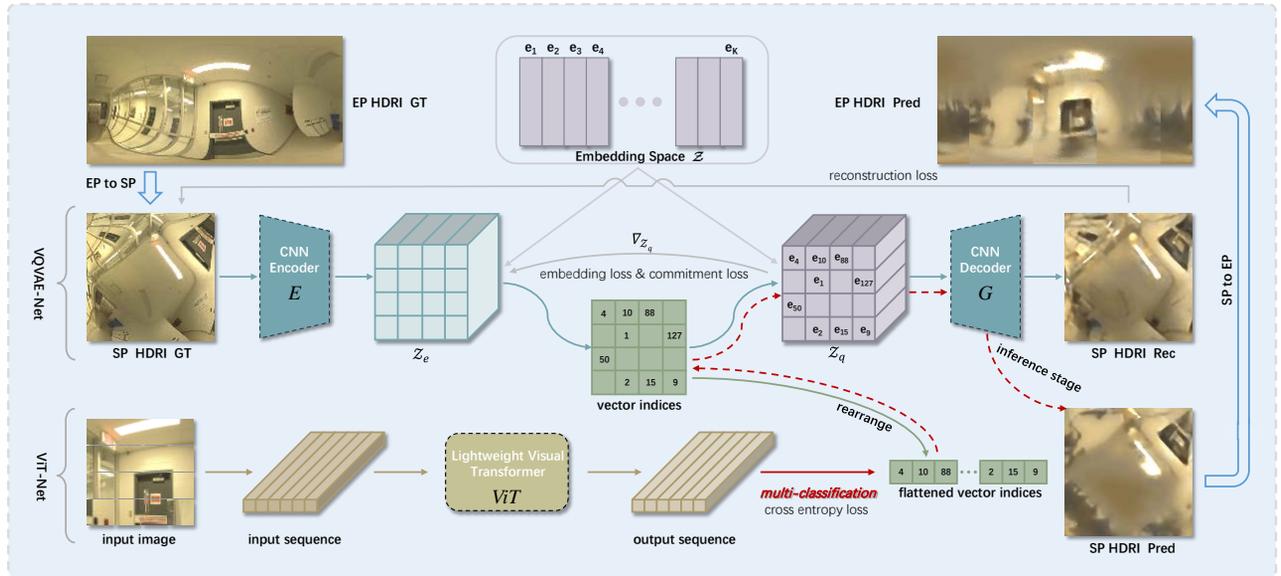

**Fig. 2**: Overview of VQT-Light. Firstly, we extract discrete features from HDR illumination map based on VQVAE architecture (VQVAE-Net), which helps to obtain the discrete latent embedding vectors ($z_q$) as well as vector indices in embedding space ($\mathcal{Z}$). Secondly, we also design a lightweight vision transformer (ViT-Net) to predict HDR light map. The ViT-Net is optimized through minimizing cross entropy loss between its output sequence and the flattened vector indices from the trained VQVAE-Net. During the testing phase, taking a single image as input, we can obtain a list of predicted vector indices by the ViT-Net, then a group of predicted latent embedding vectors, indexed by the predicted indices, are feed into the frozen VQVAE-Net decoder, the decoder produces a prediction result of illumination represented as an HDR illumination map.

## 3 Method

### 3.1 Overview of Approach

An overview of our approach is depicted in Figure 2. The proposed framework consists of two modules: feature extraction and lighting estimation. The goal of feature extraction module is to acquire discrete features of

HDR illumination map. To achieve this goal, we trained a network (VQVAE-Net) based on VQVAE architecture, which is shown in the middle. With the trained feature extraction module, we also design a lightweight vision transformer network (ViT-Net) to establish the mapping between the input image and its corresponding HDR illumination map, which is shown at the bottom.

Before feeding the training data into the framework, we follow our previous method [24] to preprocess the training dataset. Specifically, we need to perform a projective transformation on HDR illumination map $I \in \mathbb{R}^{H \times W \times 3}$, $(W = 2 \cdot H)$. By transforming the equirectangular projection (EP) to the square equal-area projection (SP) [27], we can obtain a new representation of illumination map $I' \in \mathbb{R}^{H \times H \times 3}$. To do so, the shape of illumination map changes from rectangle to square, which has been proven that it helps to produce better prediction results and build more lightweight model [24].

During the training stage, taking a square HDR illumination map as input, we extract discrete features by the VQVAE-Net, in this way, we can obtain the corresponding discrete latent embedding vectors $z_q \in \mathbb{R}^{h \times w \times D}$ as well as vector indices $C = \{i \mid i \in \{0, 1, 2, \cdots, K-1\}\}$, $|C| = h \cdot w$ from the learnt embedding space (codebook) $\mathcal{Z} \in \mathbb{R}^{K \times D}$, where $h$, $w$ represent height and width of $z_q$, $D$ and $K$ are the dimensionality and total number of embeddings in the codebook. Then, the training of the ViT-Net is treated as solving a multi-classification problem, we optimize the ViT-Net through minimizing cross entropy loss between its output sequence and the flattened vector indices $C$ from the VQVAE-Net.

In the inference stage, given a single image as input, the ViT-Net as a classifier helps to predict a sequence of indices of embeddings, then we can get the predicted embeddings by indexing the predicted indices. Finally, the predicted embeddings are feed into the frozen VQVAE-Net decoder to output the prediction result of illumination.

### 3.2 VQVAE-Net for Feature Extraction

Our previous work [24] extracts continuous latent code with a small dimensionality from illumination map, leading to many texture details are lost. In this work, with the purpose of improving this problem, we design a new feature extraction module by using VQVAE architecture, which helps to circumvent issues of "posterior collapse" and control the dimensionality reduction of features.

The details of the VQVAE-Net are shown in Table 1, the network includes an encoder ($E$), a codebook ($\mathcal{Z}$) and a decoder ($G$). The structure of the encoder and the decoder is simple, no normalization layers are used, and ReLU activation function is employed after each layer (except the last layer).

In the encoder, given an illumination map $I' \in \mathbb{R}^{128 \times 128 \times 3}$ as input, there are two convolutional layers with a kernel size of 4×4 and a stride of 2, then followed by two residual blocks [28]. The above four layers constitute the encoder for downsampling of features. The encoder's output is the continuous latent vectors $z_e \in \mathbb{R}^{32 \times 32 \times 256}$.

**Table 1**: The structure of the VQVAE-Net. It consists of an encoder, a codebook and a decoder. The input and output are square HDR illumination map with a shape of 3×128×128.

|  | Layer | Shape |
| --- | --- | --- |
| Input |  | 3×128×128 |
| Encoder | Conv2d | 128×64×64 |
|  | Conv2d | 256×32×32 |
|  | Residual Block | 256×32×32 |
|  | Residual Block | 256×32×32 |
| Codebook | Embedding |  |
| Decoder | ConvTranspose2d | 128×64×64 |
|  | Residual Block | 128×64×64 |
|  | ConvTranspose2d | 32×128×128 |
|  | Residual Block | 32×128×128 |
|  | Conv2d | 3×128×128 |
| Output |  | 3×128×128 |

Between the encoder and the decoder, there is a learnable codebook $\mathcal{Z} \in \mathbb{R}^{K \times D}$, in which the total number of embedding vectors $K$ is set to 128, and the dimensionality $D$ is set to 256.

In the decoder, the input is discrete latent embedding vectors $z_q \in \mathbb{R}^{32 \times 32 \times 256}$, which obtained from $z_e$ and $\mathcal{Z}$ by vector quantization. The operation of quantization $\mathbf{q}(\cdot)$ of each latent vector $z_e^{ij} \in \mathbb{R}^{256}$ onto its closest codebook entry $e_k \in \mathcal{Z}$ is given by

$$z_q = \mathbf{q}(z_e) := \left( \arg\min_{e_k \in \mathcal{Z}} \left\| z_e^{ij} - e_k \right\| \right) \in \mathbb{R}^{32 \times 32 \times 256} \quad (1)$$

$$(k = 0, 1 \cdots, 127;\ i, j = 0, 1 \cdots, 31)$$

To upsample feature maps in the decoder, we use two transposed convolutional layers with a kernel size of 4×4 and a stride of 2, there are also two residual blocks with the same structure as in the encoder. Finally, we use a convolutional layer with a kernel size of 1×1 to output the reconstruction result $\hat{I}' \in \mathbb{R}^{128 \times 128 \times 3}$. The reconstruction $\hat{I}' \approx I'$ is given by

$$\hat{I}' = G(z_q) = G(\mathbf{q}(E(I'))) \quad (2)$$

The loss functions used by the standard VQVAE network [25] have three components, which include a reconstruction loss, an embedding loss and a commitment loss. As for our VQVAE-Net, it has a little different with [25] when computing the reconstruction loss. We use a logarithmic transform on HDR illumination map to improve the stability of training. The entire loss function is given by

$$\mathcal{L}_{\text{vqvae-net}} = \left\| \log(\hat{I}'+1) - \log(I'+1) \right\|_2 + \left\| \text{sg}[z_e] - z_q \right\|_2 + \beta \left\| \text{sg}[z_q] - z_e \right\|_2 \quad (3)$$

where the first term is the reconstruction loss, $\text{sg}[\cdot]$ denotes the stop-gradient operation, the second term and the third term are embedding loss and commitment loss respectively. The weight of commitment loss $\beta$ is set to 0.25.

### 3.3 ViT-Net for Lighting Estimation

In order to more accurately predict illumination outside the field of view, we need to capture global context and dependencies from input image. CNNs are good at learning local features, whereas ViTs excel at learning global features, thus we design the lighting estimation module (ViT-Net) based on vision transformer.

The details of the ViT-Net are shown in Figure 3, comparing to the original structure of ViT [26], we have made some changes to predict illumination. Looking up from below, by cutting the input image with a shape of 256×256×3 into small patches with a shape of 8×8×3, there are $N = 1024$ patches $x \in \mathbb{R}^{N \times D_p}$ to be feed into a liner projection, where $D_p = 8 \times 8 \times 3$. Through the liner projection, the patches $x$ are projected to the patch embeddings with a shape of 1024×256. As we all know, the original ViT network concatenates patch embeddings, position embeddings and an extra learnable class embedding into a new data sequence before feeding to the "Transformer Encoder". However, we abandon using the extra learnable class embedding and only concatenate patch embeddings and position embeddings. The reason is that we are going to execute a multiclass classification task, single class embedding will not work. More importantly, we have demonstrated that using multiple extra class embeddings cannot lead to better result (Section 4.6).

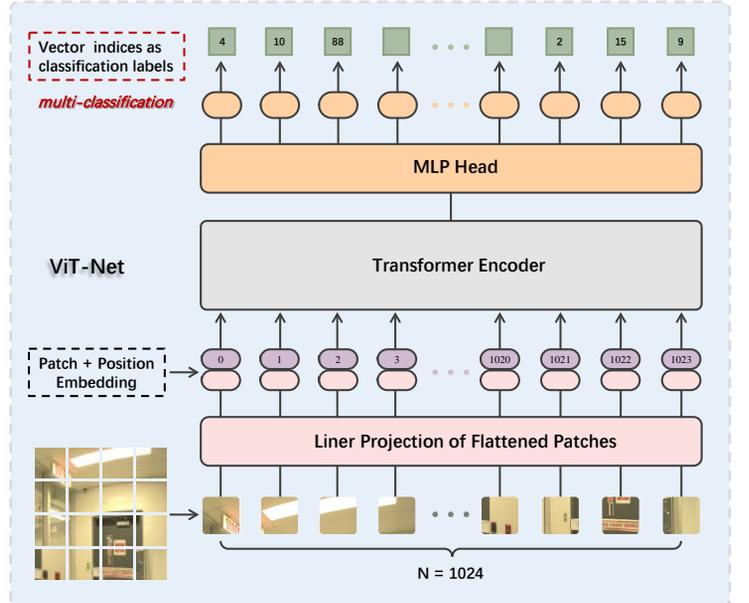

**Fig. 3**: The structure of the ViT-Net. We design the lighting estimation module based on vision transformer. Compare with the original ViT network, we discard the extra "classification token" before entering the transformer encoder. And the training of our ViT is supervised by vector indices, which obtained from the trained VQVAE-Net.

When entering the "Transformer Encoder", the depth of it is only 6, and each multi-head attention block in it has 8 attention heads. These settings help to limit the computational expense of the ViT-Net. The output of the "Transformer Encoder" is then feed into the "MLP Head", finally, the output of the "MLP Head" is a data sequence $y \in \mathbb{R}^{N \times K}$ where $K$ is the number of embeddings in the VQVAE codebook, and $N$ is the number of image patches. Actually, $N$ is determined by the total count of the classification labels, that is the vector indices $C$, we have $|C| = 1024$ so as to $N$ is set to 1024. The above process can be denoted by $y = \text{ViT-Net}(x)$.

When we use the vector indices as classification labels to supervise the training of the ViT-Net. A cross entropy loss function is applied, which is given by

$$\mathcal{L}_{\text{vit-net}} = -\frac{1}{N} \sum_{n=1}^{N} \log\left(\frac{\exp(y_{i,c})}{\sum_{j}^{K} \exp(y_{i,j})}\right) \tag{4}$$

where $N = 1024$, $K = 128$, $c \in C = \{i \mid i \in \{0, 1, 2, \cdots, K-1\}\}$.

By means of the ViT-Net, we can get the predicted indices $\hat{C}$, then find the predicted discrete embeddings $\hat{z}_q$ from the learnt codebook of the VQVAE-Net, $\hat{z}_q = \{e_j \in \mathcal{Z} \mid j \in \hat{C}\}$. At last, $\hat{z}_q$ are feed into the decoder of the VQVAE-Net to produce an HDR illumination map, which is given by

$$\hat{I}' = G(\hat{z}_q) \tag{5}$$

## 4 Experiments and Results

### 4.1 Datasets

To build the training and evaluation data for VQT-Light, we use the Laval Indoor Dataset [9] that consists of 2,233 HDR illumination maps taken in varieties of indoor scenes. Similar to [23], we crop ten images with limited field of views from each illumination map which produces 22,330 data pairs. For each of the 22,330 images, the same image warping operation in [9] is applied. Following the train/test split of EverLight [23], we use 2,009 HDR illumination maps for training and 224 illumination maps for testing.

In our experiments, we also need to transform the equirectangular projection to the square equal-area projection [27] for all of the 22,330 HDR illumination maps, then we can acquire the data pairs $\{I, I', P\}$ as used, where $I \in \mathbb{R}^{128 \times 256 \times 3}$ represents HDR illumination map with the equirectangular projection, $I' \in \mathbb{R}^{128 \times 128 \times 3}$ represents HDR illumination map with the square equal-area projection, and $P \in \mathbb{R}^{256 \times 256 \times 3}$ represents the cropped LDR image from its corresponding illumination map.

### 4.2 Implementation Details.

Our model is implemented by the PyTorch framework. We employ a single NVIDIA TITAN XP GPU in the experiments. The VQVAE-Net is trained for 20 epochs, and optimized by Adam with a batch size of 16 and an initial learning rate of 5e-4. We also use an exponential learning rate decay mechanism with the decay parameter gamma=0.92. As for the ViT-Net, it is trained for 25 epochs, and also optimized by Adam with a batch size of 16 and an initial learning rate of 1e-3, we employ a multi-step learning rate decay mechanism with the decay parameters milestones = [15, 20] and gamma=0.1.

### 4.3 Evaluation Metrics

Similar to EMLight [18] and StyleLight [22], we employ three spheres with different materials for evaluation, including mirror, glossy and diffuse as illustrated in Figure 4. We render the three spheres by Blender with the predicted HDR illumination map and its corresponding ground-truth. In the process of rendering, we use the "Cycles" render engine, which is a physically-based path tracer in Blender.

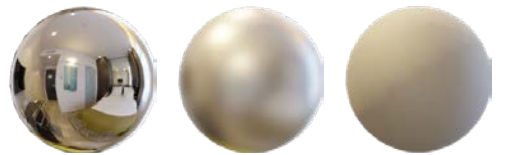

**Fig. 4**: Three spheres with different materials: mirror, glossy, and diffuse.

The evaluation metrics include Root Mean Square Error (RMSE), Structural Similarity (SSIM), Angular Error and Learned Perceptual Image Patch Similarity (LPIPS), which uses VGG16 [29] as the perceptual network. Theses metrics have been widely adopted on rendered images in the evaluation of illumination prediction. Note each rendered image contains three spheres with different materials at the same time. In addition, we also use Fréchet Inception Distance (FID) [30] to measure the distribution similarity between the estimated illumination maps and the ground-truth.

## 4.4 Qualitative Evaluation

In this section, we compare VQT-Light with several state-of-the-art methods qualitatively, which include four regression-based methods Gardner'17 [9], Gardner'19 [12], Garon'19 [11], Xie'24 [24] and four generation-based methods Lighthouse [15], EMLight [18], StyleLight [22], EverLight [23]. As depicted in Figure 5, taking a single image as input, we present the visualization of the predicted HDR illumination map and a rendered image composed of three spheres with different materials (Fig 4) on a diffuse plane.

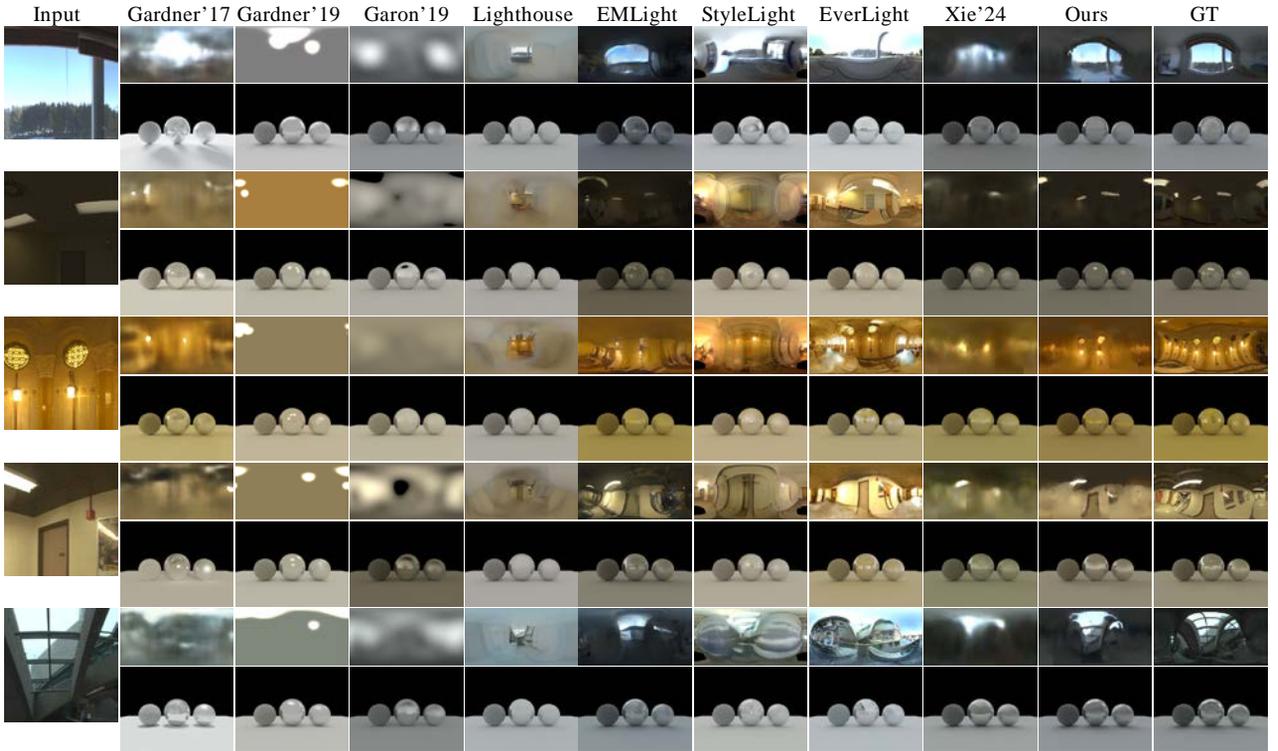

**Fig. 5**: Visual comparison of VQT-Light with state-of-the-art methods. From a given input image (left), we show the estimated illumination map (first row) and the rendered image (second row) composed of three spheres (diffuse, mirror, glossy) on a diffuse plane.

Gardner'17 regresses and predicts illumination map directly by CNNs, which fails to recover texture details of illumination map. Gardner'19 regresses spherical Gaussian parameters, which tends to focus on strong light sources and produce inaccurate ambient lighting. Garon'19 regresses the 5th-order spherical harmonic (SH) coefficients while it struggles to recover accurate high-frequency information of light. As we can see, the light maps predicted by these methods are not visually similar to the ground truth, the rendered images also have gaps with the reference rendered images in terms of shading, shadows and reflections. Besides, Lighthouse can only output LDR illumination map and may lead to a large bias on rendering. EMLight generates illumination map under the guidance of predicted SG parameters, which are difficult to train because of the unstable regression. StyleLight and EverLight are able to generate illumination map with high-quality textures, but the textures of predicted light map may be quite different from the ground truth, resulting in the rendered images are not similar to the reference images. Xie'24 regresses continuous latent codes of illumination map and outputs a blurred illumination map, it also has the problem of losing many texture details. In contrast, VQT-Light can predict an illumination map with richer texture, especially in areas inside the field of view. The appearance of three spheres in the rendered images are quite similar to the reference.

### 4.5 Quantitative Evaluation

Table 2: Quantitative comparison with several state-of-the-art methods. The evaluation metrics include the widely used RMSE, SSIM, Angular Error and LPIPS of the rendered images. Besides, FID is also used to measure the distribution similarity between the estimated illumination maps and the ground-truth. Bold font indicates the best result.

|  | RMSE↓ | SSIM↑ | Angular Error↓ | LPIPS↓ | FID↓ | Projection |
| --- | --- | --- | --- | --- | --- | --- |
| Gardner'17[9] | 0.0773 | 0.9415 | 2.082 | 0.1040 | 309 | EP |
| Gardner'19[12] | 0.0525 | 0.9629 | 2.018 | 0.0599 | 356 | EP |
| Garon'19[11] | 0.0515 | 0.9614 | 2.760 | 0.0628 | 352 | EP |
| Lighthouse[15] | 0.0432 | 0.9710 | 3.435 | 0.0687 | 219 | EP |
| EMLight[18] | 0.0435 | 0.9687 | 1.656 | 0.0518 | 221 | EP |
| StyleLight[22] | 0.0499 | 0.9640 | 1.967 | 0.0533 | 148 | EP |
| EverLight[23] | 0.0441 | 0.9675 | 1.817 | 0.0501 | **135** | EP |
| Xie'24[24] | 0.0359 | 0.9809 | 1.377 | 0.0455 | 329 | SP |
| VQT-Light(ours) | **0.0324** | **0.9880** | **1.356** | **0.0400** | 256 | SP |

We compare VQT-Light with several state-of-the-art methods quantitatively. As shown in Table 2, we can observe that our method outperforms the compared methods on all of metrics for rendered images (the first four items). Furthermore, generation-based methods outperform regression-based methods on the metrics for rendered images, but expect for Xie'24 and VQT-Light. The regression-based methods, such as Gardner'19 and Garon'19, cannot recover accurate low- or high-frequency information of light, the predicted illumination maps lose many texture details. The generation-based methods, such as StyleLight and EverLight, can generate more realistic illumination map, but they have a common disadvantage that the content of generated illumination map may differ from the ground truth significantly, especially in areas outside the field of view. This disadvantage causes an adverse effect on the metrics for rendered images. For Xie'24 and VQT-Light, we believe the use of new projection scheme plays an important role, resulting in better performance than other methods. Comparing to Xie'24, VQT-Light combines neural discrete representation and global self-attention mechanism, which helps to recover more textures of illumination map and predict the light sources outside the field of view.

On the metric of FID, VQT-Light is inferior to the generation-based methods, EverLight obtains the best score. In this work, the light map textures are still difficult to accurately predict in areas outside the field of view, which causes negative influence on the distribution similarity between the predicted light maps and the ground-truth. Nevertheless, VQT-Light acquires the best score of FID among all the regression-based methods.

Moreover, our approach does not require any time-consuming computation, and takes about 0.025 seconds to perform once inference. Using the same experimental platform, the inference time of other methods is listed in Table 3. In summary, Garon'19, Xie'24 and VQT-Light can predict illumination in real-time, their running speeds are 53FPS, 48FPS and 40FPS respectively, VQT-Light is not the fastest but achieves the most accurate estimation.

Table 3: Time cost comparison of several state-of-the-art methods in once inference. Garon'19, Xie'24 and VQV-Light reach real-time running speed.

|  | Gardner'17 | Gardner'19 | Garon'19 | Lighthouse | EMLight | StyleLight | EverLight | Xie'24 | VQT-Light (ours) |
| --- | --- | --- | --- | --- | --- | --- | --- | --- | --- |
| Time (sec) | – | – | 0.019 | – | 0.16 | 65.48 | 0.079 | 0.021 | 0.025 |

### 4.6 Ablation Studies

Using the same evaluation metrics, we further evaluate VQT-Light by developing several variants, which include 1) applying different feature resolution in the VQVAE-Net, 2) using different numbers of embeddings ($K$) in the codebook of VQVAE, 3) using CNNs rather than ViT to build the lighting estimation module, and 4) applying extra class embeddings when building the ViT-Net.

At first, we compare the reconstruction quality of HDR illumination map by applying different resolution of latent feature maps in the VQVAE-Net. We present the reconstruction loss curves in Figure 6, which shows that the higher feature resolution, the less reconstruction loss when converged. However, we have to limit the growth of feature resolution, because the length of flattened latent indices ($C$), which determined by the feature resolution, cannot be too long. Otherwise, the input/output sequence of the ViT-Net will be too long, which can severely influence network performance. Therefore, in order to achieve a balance between prediction accuracy and inference speed, we choose 32×32 as the resolution of features to be extracted.

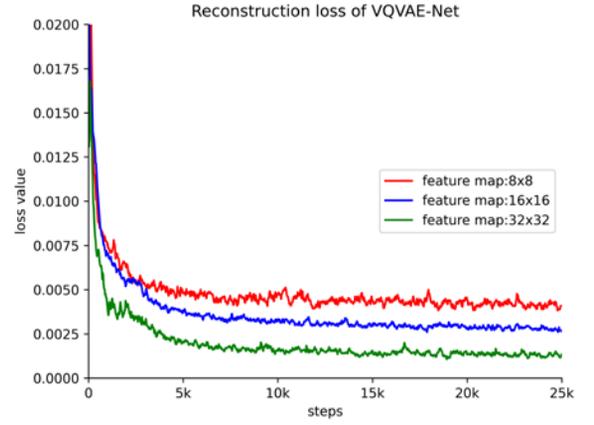

**Fig. 6**: Comparison on reconstruction loss of the VQVAE-Net by using different feature resolution.

**Table 4**: Quantitative comparison on different options quantitatively. VQV-Light uses the second option.

| | Options | RMSE↓ | SSIM↑ | Angular Error↓ | LPIPS↓ | FID↓ | Speed↑ |
|---|---|---|---|---|---|---|---|
| (1) | $K$=64+ViT w/o extra class tokens | 0.0335 | 0.9845 | 1.507 | 0.0420 | 273 | 42 |
| (2) | $K$=128+ViT w/o extra class tokens | 0.0324 | 0.9880 | **1.356** | **0.0400** | 256 | 40 |
| (3) | $K$=256+ViT w/o extra class tokens | **0.0310** | **0.9892** | 1.360 | 0.0406 | **251** | 35 |
| (4) | $K$=128+CNNs instead of ViT | 0.0385 | 0.9765 | 1.583 | 0.0472 | 330 | **60** |
| (5) | $K$=128+ViT w/ extra class tokens | 0.0339 | 0.9877 | 1.536 | 0.0442 | 282 | 37 |

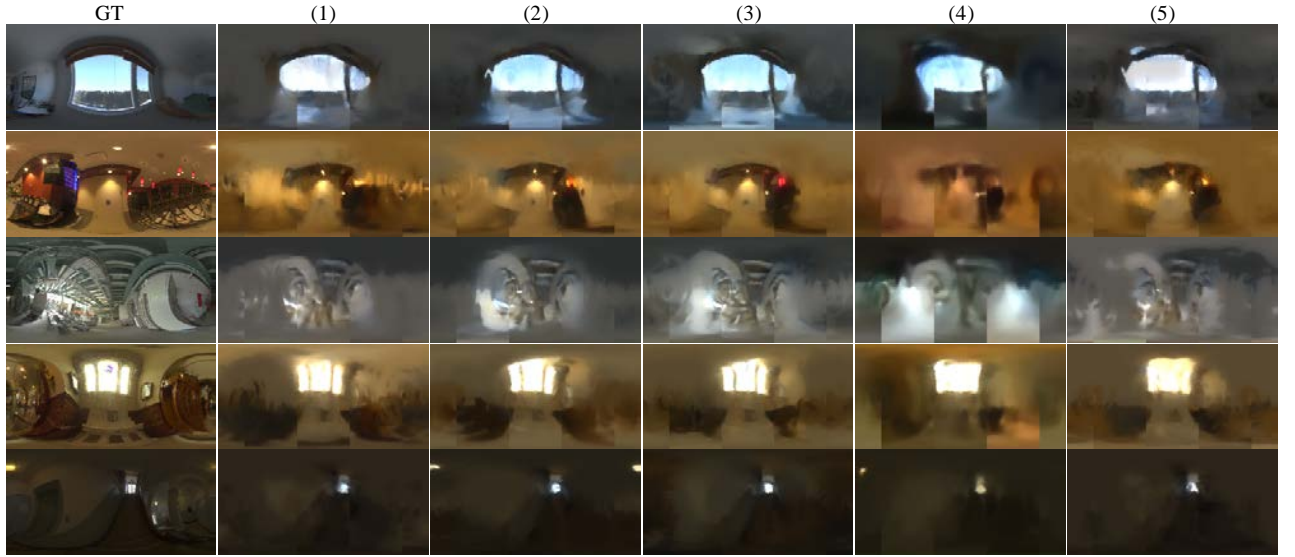

**Fig. 7**: Visual comparison on different options listed by Table 4. We visualize the illumination map predicted by each variant, the ground truth is shown in the first column. VQT-Light is shown in the third column.

Then we evaluate the metrics by using different numbers of embeddings ($K$) in the codebook of the VQVAE-Net. As Table 4 shows, when $K$=64, the inference speed reaches 42FPS, which is faster than the options of $K$=128 and 256, but the remaining metrics are not as good as the other two options. Comparing $K$=128 with $K$=256, each option has its own advantages and disadvantages. In Figure 7, we can observe that the option 2 is comparable with the option 3, and both of them outperform the option 1 in visual. Taking account into the speed requirements of practical applications, we choose $K$=128 as the number of embeddings in the codebook.

Next, we evaluate the performance by using CNNs instead of ViT to build our lighting estimation module, that is, comparing option 4 with option 2, we keep the total number of network parameters close to each other, the metrics in Table 4 indicate that using CNNs performs worse than using ViT, though the running speed of option 4 is faster. Comparing with CNNs, we believe that the capability of ViT, which good at capturing long-range dependencies and global features in an image, plays a key role in our framework.

Finally, we evaluate the performance by applying extra class embeddings when building the ViT-Net. We know that the original vision transformer uses an extra class token to predict a class label. In our multi-classification task, using additional class tokens will increase the length of data sequences of ViT, to which the transformer encoder is sensitive. Too long data sequence is harmful to the training of the ViT-Net. Table 4 and Figure 7 show that the option 5 performs not well both quantitatively and qualitatively. Therefore, VQT-Light gives up applying extra class embeddings.

## 5 Discussion

In this work, there are some limitations need to discuss. Comparing with other regression-based methods, although we have improved the texture fidelity of illumination map to some extent, there are still a lot of texture details missing in areas outside the field of view as shown in Figure 8. This problem is caused by the fact that lighting estimation from a single image is an under-constrained problem. While using a generative model is able to generate high-quality textures, the generated textures may be significantly different from the ground truth even the model is conditioned by the input image.

To solve this problem, we believe that additional input data is required. Therefore, we consider to relax the limitation of input in future work, such as using a pair of photos taken by the front and rear cameras of a smartphone respectively. In this way, we can further optimize the prediction of illumination map, and improve the realistic rendering of virtual objects.

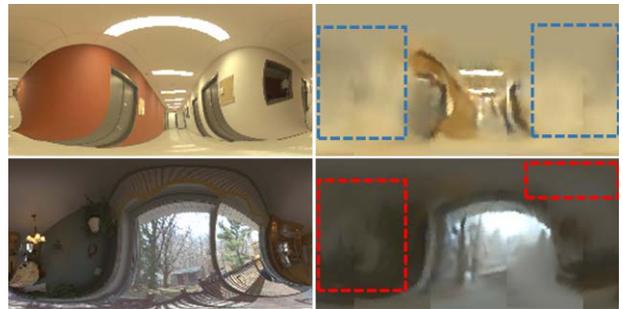

**Fig. 8**: Examples of failure to predict textures of illumination map in areas outside the field of view, which marked with blue/red boxes.

## 6 Conclusion

In order to predict illumination map with richer texture in real-time, and avoid the trap of focusing on generating high-quality textures but sacrificing the content fidelity of light map, we propose a novel framework that combining the advantages of both VQVAE and ViT architecture, and translate the under-constrained lighting estimation problem to a multiclass classification task. When designing the VQVAE-Net, we explore appropriate resolution of features and suitable quantity of embedding vectors in the codebook. When designing the ViT-Net, we remove redundant parts and find out suitable structure of it. Comparing with regression-based state-of-the-art methods, experiments show that VQT-Light produces qualitatively and quantitatively superior results. Comparing with state-of-the-art generation-based methods, VQT-Light achieves more accurate results on rendered images, but poor performance of FID owing to the lack of texture in some areas. In summary, VQT-Light improves RMSE, SSIM, Angular Error and LPIPS by 9.7%, 0.7%, 1.5% and 12.1% respectively, and achieves a running speed with 40FPS. In future work, we plan to moderately relax the input constraints according to practical applications to predict illumination maps with higher quality.